\documentclass{article}
\usepackage{acra}

\usepackage{graphicx}%
\usepackage{multirow}%
\usepackage{amsmath,amssymb,amsfonts}%
\usepackage{mathrsfs}%
\usepackage{xcolor}%
\usepackage{textcomp}%
\usepackage{manyfoot}%
\usepackage{booktabs}%
\usepackage{algorithm}%
\usepackage{algorithmicx}%
\usepackage{algpseudocode}%
\usepackage{listings}%
\usepackage{subfig}
\usepackage{graphicx}
\usepackage{subcaption}
\usepackage{url}
\title{Influence of Operator Expertise on Robot Supervision and Intervention}
\author{Yanran Jiang$^{1,}$$^{*}$, Pavan Sikka$^{1}$, Leimin Tian$^{1}$, Dana Kuli\'c$^{1, 2}$ and C\'ecile Paris$^{1}$ \\ $^{1}$Data 61, CSIRO, 
        Australia, $^{2}$Faculty of Engineering,  
        Monash University, Australia \\ 
$^{*}$yanran.jiang@csiro.au}

\begin{document}

\maketitle

\begin{abstract}
As autonomous robots are increasingly deployed in complex and high-risk environments, they are supervised by users with diverse levels of robotics expertise. As the diversity of the user population increases, it is important to understand how users with different expertise levels approach the supervision task and how this impacts performance of the human-robot team. This exploratory study investigates how operators with varying expertise levels perceive information and make intervention decisions when supervising a remote robot. We conducted a user study (\(N=27\)) where participants supervised a robot autonomously exploring four unknown tunnel environments in a simulator, and provided waypoints to intervene when they believed the robot had encountered difficulties. By analyzing the interaction data and questionnaire responses, we identify differing patterns in intervention timing and decision-making strategies across novice, intermediate, and expert users. 
\end{abstract}

%%%%%%%%%%%%%%%%%%%%%%%%%%%%%%%%%%%%%%%%%%%%%%%%%%%%%%%%%%%%%%%%%%%%%%%%%%%%%%%%
\section{INTRODUCTION}\label{sec1}

The integration of semi-autonomous remote robots into industrial, service, and field environments offers numerous benefits such as enhanced efficiency, reduced human labor, and increased safety~\cite{ajoudani2018progress}. However, the success of these robotic systems often depends on the ability of human operators to monitor and intervene when necessary~\cite{othman2023human,hopko2022human}. This is especially crucial in complex environments where the semi-autonomous remote robot may encounter situations that challenge its autonomous capabilities, leading to suboptimal performance or failures~\cite{wijayathunga2023challenges}. One such complex environment is search and rescue, where a robot must deal with varying terrain, obstacles, and other dynamic factors, requiring remote human supervision~\cite{riley2004hunt,nourbakhsh2005human}. 

Human supervision of robots can involve operators with varying levels of knowledge and experience of robot capabilities or task understanding, i.e., operator's expertise~\cite{cunha2022exploring,conlon2022m}. Comprehending how operators of varying expertise levels perceive information and make decisions is key to developing effective human-robot teaming (HRT) with shared autonomy~\cite{lewis2013adjustable}. By equipping robots with adaptive autonomy, these systems can adjust their behavior based on the operator's expertise, providing more assistance to novices while allowing experts greater control~\cite{licardo2024intelligent,sfair2020human}. To appropriately adapt to the operator's expertise, robots need to be able to infer the operator's level of expertise based on their interactions. However, to date there is limited research on how intervention strategies vary with the expertise levels of operators in remote HRT.

We conducted a user study recruiting participants with different expertise levels in a remote HRT task to characterize the strategies of user interventions, how they vary as a function of user expertise, and their impact on task performance. The HRT mission was generated in a custom simulator environment, using the same robot autonomy algorithms, data visualization, operator control interface, and realistic mission maps obtained from field deployment of the CSIRO Data61 team during the DARPA SubT challenge~\cite{kottege2023heterogeneous}. We aim to understand how operator expertise influences intervention behavior, providing insights on enhancing robot collaboration with users of diverse expertise levels. Our contributions are:

\begin{itemize}
    \item We provide an empirical analysis of intervention behaviours (timing and waypoint placement) across operator expertise levels in remote human–robot supervision tasks.
   
    \item Post-simulation analysis evaluates intervention efficiency, showing how expertise influences subsequent robot performance.

    \item We assess diverse operators' perceptions and preferences in remote supervision of robots, demonstrating the need for both human-initiated intervention and robot-initiated help requests.

\end{itemize}

%This paper is structured as follows: we first discuss related work on operator expertise in human-robot interaction and the limitations of adaptive systems. Next, we describe the methodology of our user study, including participant recruitment, experimental setup, and data collection procedures. We then present the results of our study, followed by a discussion of the findings and their implications. 

\section{Related Work}

%In this section, we review the related work analysing the role of human operators in HRT and the limitations of current adaptive systems in responding to varying levels of operator expertise.

\subsection{The role of human supervisors in human-robot teaming}
Human-robot teaming (HRT) research focuses on improving interaction and performance in complex, high-risk environments such as smart farming~\cite{vasconez2019human}, search and rescue~\cite{nourbakhsh2005human}, and space exploration~\cite{hambuchen2021review}. Human operators contribute significantly to HRT by providing foresight, contextual awareness, and advanced prioritization, ensuring that critical tasks are addressed during missions~\cite{chen2014human}. Operators typically assess the robot’s state, integrate visual information, and decide on the next steps to enhance task execution~\cite{burke2004moonlight}. Where communication allows, they intervene through guidance or teleoperation to prevent critical incidents, such as robots becoming stuck or task inefficiency~\cite{scholtz2004evaluation}.

While operators can significantly enhance mission outcomes, they can also introduce errors and interruptions that affect performance~\cite{sheridan2016human}. Sheridan~\cite{sheridan2016human} highlighted supervisory errors that undermine robot performance, while Wang et al.~\cite{wang2016trust} examined how task-switching increases errors and delays. Understanding when and how to best utilize operator intervention is crucial for optimizing human-robot collaboration. Robinson et al.~\cite{robinson2022human} evaluated the impact of human operators on mission success and team performance in search and rescue missions. By comparing human-robot teams to fully autonomous teams in real-world scenarios, they identified scenarios where operator interventions were advantageous, such as redirecting robots to better cover ground or navigate challenging terrain. They also identified occasions where human intervention was disadvantageous or unneeded. However, this study was limited to expert operators. To develop more adaptive systems and facilitate wider adoption of HRT, it is critical to consider operators with different experience levels.

\subsection{Robot adaptation to human expertise levels}

Identifying and adapting to varying operator skill levels is essential for enhancing HRT~\cite{storms2014predicting}. Conlon
et al.~\cite{conlon2022m} found that robots communicating task proficiency without considering operator expertise led to trust and performance issues, especially in novices who ignored confidence levels. Their work highlights the difficulties novice operators face in controlling robots and reveals a significant gap: robots typically assess task feasibility based solely on their own performance metrics, which can lead to overconfidence in novices~\cite{conlon2022m}. This emphasizes the need for robots to better understand human expertise in HRT.

Most existing works classify novice and expert users with a pre-assessment of operator skill~\cite{lewis2013adjustable,sfair2020human}. %For instance,~\cite{villani2020inclusive} introduced an adaptive automation framework, called INCLUSIVE, that assists the operator during working tasks. It contains a module that adapts behavior based on the operator’s limitations, improving usability and performance. They determined initial human capabilities before interaction by considering constitutional user characteristics and a questionnaire that inquired about the user’s general computer skills, work experience, experience with the machine, and the presence of visual impairments, according to the guidelines provided by~\cite{villani2019measurement}. However, relying solely on pre-assessments to classify operators as experts or novices may not capture the full picture of their actual performance~\cite{li2023proactive}. 
However, such static classifications may not fully reflect actual performance~\cite{li2023proactive}. To address this, some studies dynamically adapt to user expertise based on real-time cues such as intent and task performance across manipulation and feedback-driven learning~\cite{lewis2013adjustable,liang2024learning}. For instance, Lewis et al.~\cite{lewis2013adjustable} employs an adjustable autonomy approach that classifies expertise using k-means clustering on teleoperation features like task completion time and command frequency. The system then adjusts autonomy to support weaker areas, enhancing performance. However, it does not account for increasing expertise as a user learns over time.

%Unlike traditional static assessments and predefined rules, several studies evaluate and adapt to human performance during the task itself based on extrinsic cues such as intent or performance across diverse tasks such as manipulation and feedback-driven learning~\cite{acharya2018inference,lewis2013adjustable,liang2024learning}. For example,~\cite{lewis2013adjustable} presents an adjustable autonomy approach for multi-robot manipulation tasks, which adapts to differences in operator expertise. It models human expertise across three key tasks: navigation, manipulation, and coordination. The system collects a short teleoperation sequence, analyzing features such as task completion time and command frequency, to classify users as experts or novices in each area by applying k-means clustering. Based on these classifications, the system adjusts robot autonomy to assist the operator in weaker areas, thereby improving overall performance. However, the progression of learning over time is not considered. As familiarity with the task increases, the user's expertise level may change.

User expertise is often considered in robot learning from human input. For instance, Liang et al.~\cite{liang2024learning} use large language models (LLMs) to learn from iterative human feedback, classifying top-users based on the number of interactions needed to teach the robot. However, this approach risks conflating user skill with low standards, as users both guide and assess the robot. This highlights the need for a refined expertise assessment that considers both instruction efficiency and final task quality. In the field of robot learning from demonstration, participants' performance is often evaluated using metrics focused on final task outcomes, such as task success, time spent on demonstrations, consistency in object handling, and the number of help requests~\cite{cakmak2014teaching,sakr2020training}. These indicators reflect the effectiveness and skill of participants at task demonstration, which are only partially useful for assessing the operator expertise level in remote HRT, corresponding most closely to supervisor teleoperation.

%User expertise is often considered in robot learning from human input.  For example,~\cite{liang2024learning} use large language models (LLMs) to implement robot learning from iterative human feedback.  They differentiate between top-users and other users by the number of interaction rounds the user requires to correctly teach the robot to the user's satisfaction. However, this approach can conflate high-performance user feedback with low  user standards, as users both guide and assess the robot. This highlights the need for a more refined method of identifying expertise—one that considers not only the efficiency of user instructions but also the quality of the final task outcome. In the field of robot learning from demonstration, participants' performance is often evaluated using metrics focused on final task outcomes, such as task success, time spent on demonstrations, consistency in object handling, and the number of help requests~\cite{cakmak2014teaching, cakmak2014eliciting,sakr2020training,pais2015metrics}. These indicators reflect the effectiveness and skill of participants at task demonstration, which are only partially useful for assessing the operator expertise level in remote HRT, corresponding most closely to supervisor teleoperation.

There is a lack of research exploring user behaviours as expertise indicators in remote HRT beyond teleoperation. This exploratory study examines how expertise influences robot supervision and task performance. We conducted a user study in a realistic simulated search and rescue scenario, analyzing participants' monitoring and intervention strategies during a tunnel exploration mission. Findings provide insights to enhance adaptive and effective HRT design.

\section{Experiment}

To investigate how operators with varying expertise levels perceive information and provide intervention in remote HRT, we conducted a user study, approved by the CSIRO Ethics Review Board. Related videos are in \url{https://yanranjiang.github.io/Influence-of-Operator-Expertise/}.

\subsection{Research Questions}
We investigate the following research questions:

\begin{itemize}
    \item \textbf{RQ1}: How does the expertise level of human operators impact their intervention behaviours?
    
    \begin{itemize}
       \item \textbf{RQ1.1}: Do operators with higher levels of expertise intervene at times closer to the pre-designed robot failures?
       \item \textbf{RQ1.2}: Do operators with higher levels of expertise provide more effective interventions, such as positioning waypoints that better help the robot recover from failures and enhance exploration outcomes?
    \end{itemize}
    \item \textbf{RQ2}: Do operators, particularly novices, prefer a robot that explicitly requests help over relying on human observation to initiate intervention?
\end{itemize}

\subsection{User Study Design}

The user study simulated an HRT mission where the aim was to explore as much of an unknown tunnel environment as possible within a time limit. The team was composed of one remote human operator and one simulated BIA5-tracked all-terrain robot (ATR). During the automated exploration phase, the robot may encounter various challenges that can result in reduced progress or complete immobilization. The key challenges the robot faces include identifying and avoiding redundant exploration of areas it has already visited, efficiently navigating through narrow tunnels, and safely traversing steep slopes (see details in Section \ref{sec:simu}). We ask the operator to supervize the robot and intervene when necessary to help the robot explore the map efficiently. Participants are expected to intervene by providing waypoints to the robot (assist in navigating challenging terrain) when they deem it necessary. Only one intervention was allowed in each exploration mission to ensure consistency and comparability across participants. The robot's actions in the simulator were pre-recorded, ensuring that each participant observed the exact same robot execution trajectory up to the point of intervention. This standardized the robot’s state and enabled systematic measurement of timing differences across expertise levels. The simulation was terminated as soon as the participant provided a waypoint. Participants' intervention time and waypoint positions were recorded for data analysis and to enable forward simulation for assessing the effectiveness of the intervention.  
%Simulated robot's path is prerecorded in rosbag and replay to participants. Participants only need to intervene once and then the corresponding scenario will be ended and then go to next one. After each scenario, participants are required to rate their intervention timeing (timely, early, or late) and give their overvations. 

\subsection{Simulation Environment}
\label{sec:simu}

We utilize the CSIRO NavStack simulator~\cite{kottege2023heterogeneous}, which accurately replicates the capabilities of the physical robot. The simulated robot was equipped with a multi-agent 3D Simultaneous Localisation and Mapping (SLAM) system~\cite{ramezani2022wildcat}, along with multiple RGB and thermal cameras and a spinning lidar. The robot can autonomously explore and construct an accurate map of an unknown area for navigation and localization of artifacts, and communicate its progress to the base station (operator interface)~\cite{robinson2022human}. 

%The robot trajectories are planned using hybrid \(A^{*} \), as detailed in~\cite{kottege2023heterogeneous,hines2021virtual}. It dynamically updates the trajectory based on the robot's current position and the nearest obstacles, adjusting the robot's speed to navigate precisely around obstacles. However, hybrid \(A^{*} \) is not guaranteed to produce a safe path. The robot may get stuck in local navigation due to several reasons: high obstacle density, dynamic obstacles, kinematic limits, and steep slopes. Additional issues, such as trajectory regeneration failures, unreachable planning nodes, excessive pitch/roll, or sensor malfunctions, can further hinder navigation. While recovery behaviors like Orientation Correction and Decollide~\cite{hines2021virtual} help mitigate these challenges, they may not always succeed, requiring intervention from the remote human operator.

%Fig.~\ref{fig:sim_GUI} shows an overview of the operator interface (simulator). 
The operator interface (simulator) is segmented into four panels (see supplementary video for details): the Executive panel (left) allows the supervisor to select a robot and manually generate tasks for it to execute, namely a sequence of waypoints to navigate to, as well as an E-Stop panel for shutting down a selected robot in an emergency; the Map panel (center) displays a simulated ATR robot in a 3D map that is generated and updated continuously using SLAM and navigation data. The robot's planned trajectory is also shown. On this 3D map, the supervisor can add waypoints to create supervisor-generated tasks for exploration, which are then added to the existing task list in the Executive panel for execution; the Robot Camera's Feed panel (top right) streams the robot's front and/or rear camera feeds; the Visualisation Manager panel (bottom right) displays status information of the robot, such as network connectivity and errors.

%\begin{figure}
    %\centering
    %\includegraphics[width=0.8\linewidth]{Fig1.1.png}
    %\caption{An overview of the user interface.}
    %\caption{User interface overview: a remote human operator can view the robot's onboard cameras (right panels), visualize the map explored with additional information such as the local cost map (middle panel), and override the robot's autonomous exploration by specifying a list of waypoints in the map for it to navigate to in the given order (left panels).}
    %\label{fig:sim_GUI}
%\end{figure}

\begin{figure*}
    \centering
    \subfloat[Map of scenario 1 \label{fig:sc1}]{
        \includegraphics[width=4cm,height=3cm]{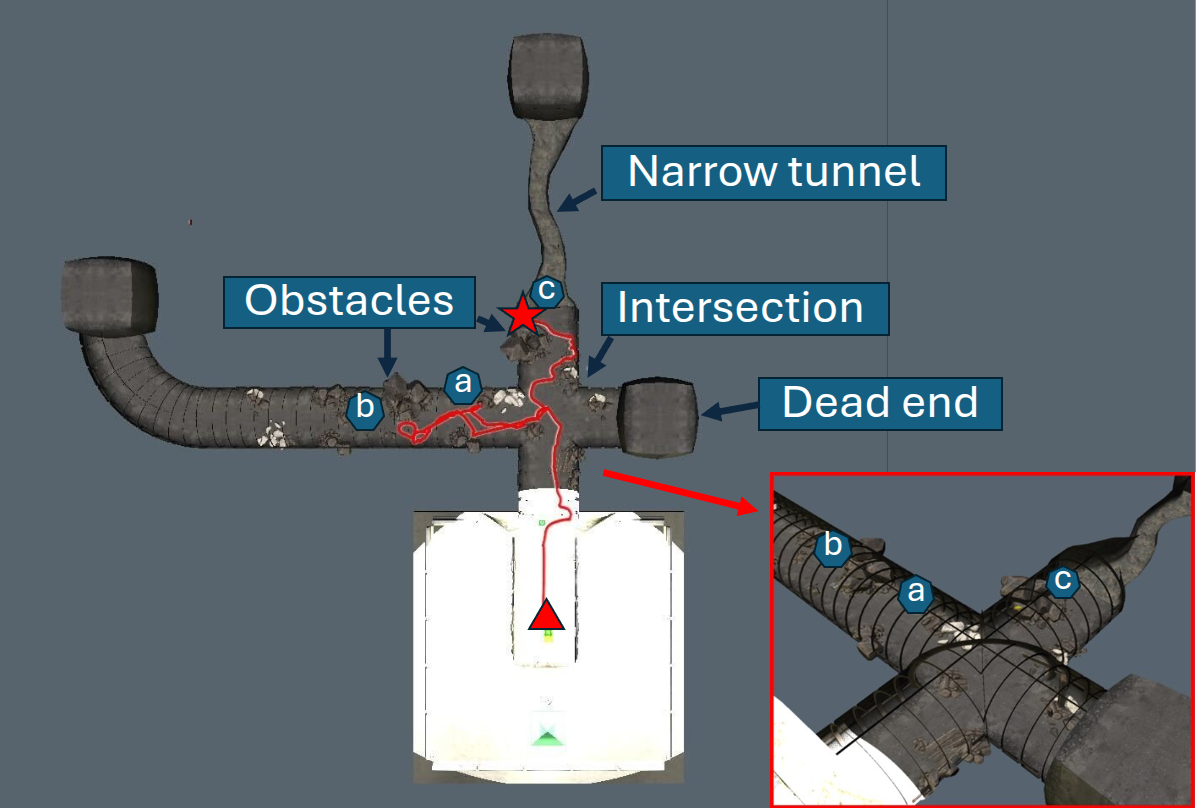}}
    \subfloat[Map of scenario 2 \label{fig:sc2}]{
        \includegraphics[width=4cm,height=3cm]{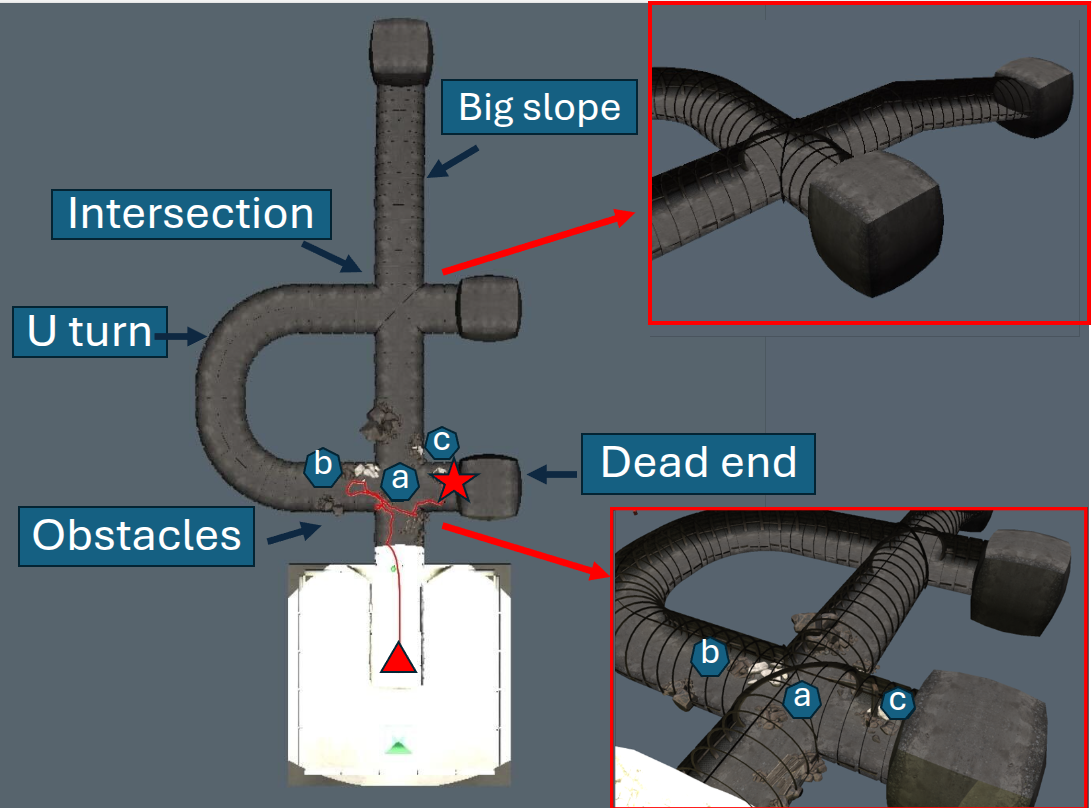}}
    \subfloat[Map of scenario 3 \label{fig:sc3}]{
        \includegraphics[width=4cm,height=3cm]{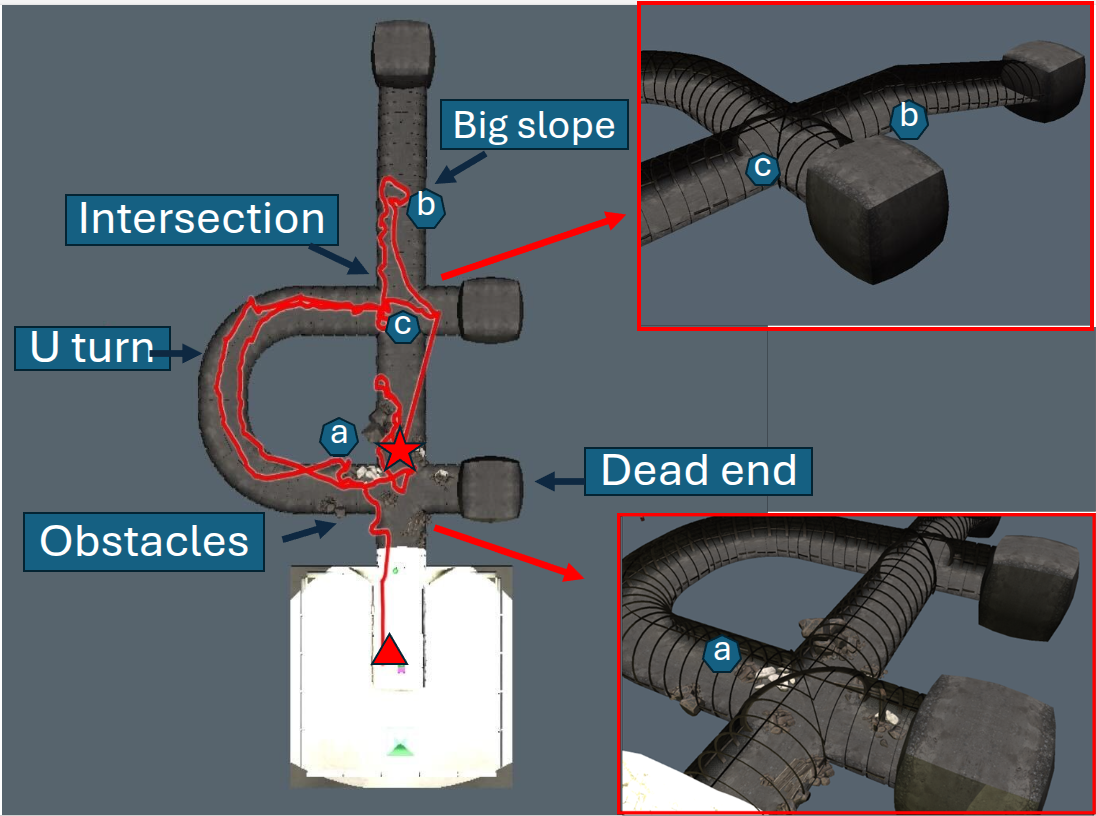}}
    \subfloat[Map of scenario 4 \label{fig:sc4}]{
        \includegraphics[width=4cm,height=3cm]{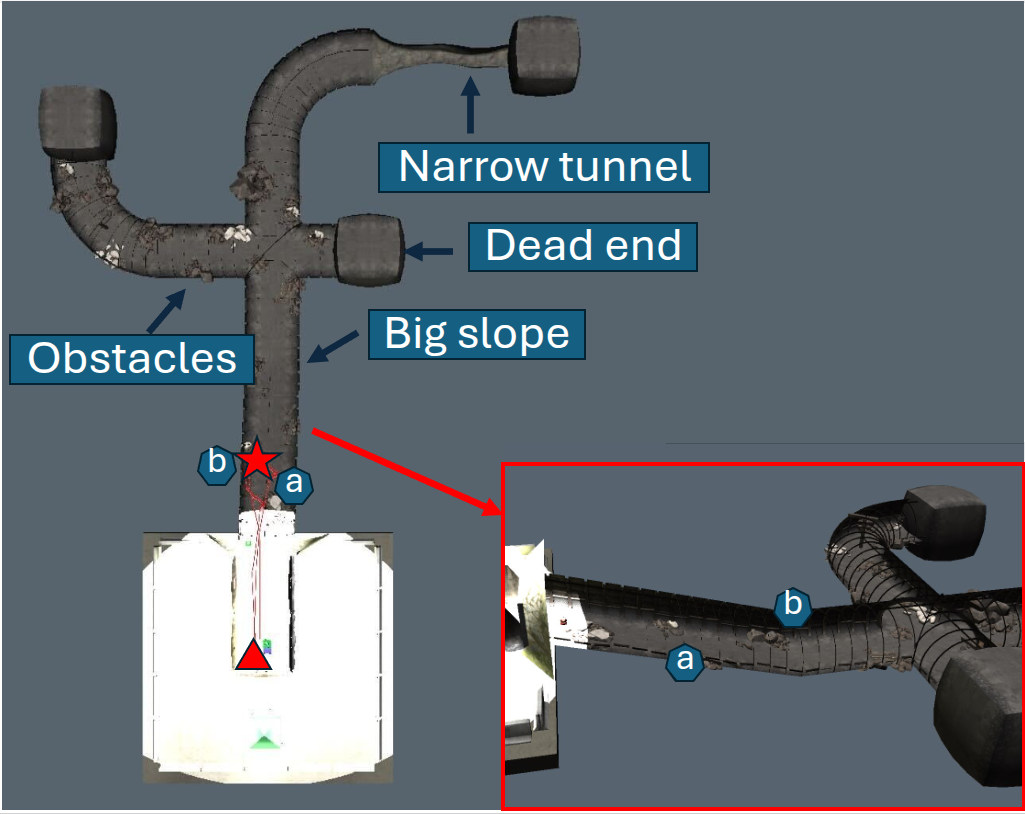}}

    \caption{Four scenarios with annotated challenging tunnel maps. The robot's trajectory (red line) is extracted from recorded data. Red triangle marks trajectory start and red star marks the end. Dark blue annotations indicate points of robot delays. Predefined failure points, requiring human intervention, are marked at location c in scenarios 1-3 and b in scenario 4.}

    \label{fig:map} 
  
\end{figure*}

\subsection{Simulator Scenario Design}
\label{sec:sc_design}
We designed four simulator maps based on real-world challenges~\cite{kottege2023heterogeneous}, incorporating steep slopes, constrained tunnels, intersections, and sharp turns. These elements were arranged in feasible configurations to create diverse scenarios (Fig.~\ref{fig:map}). The maps highlight key locations where robot progress is impeded.
Each scenario introduces specific challenges, with intentionally designed robot failures at location c for Scenarios 1-3 and b for scenario 4:

\begin{itemize}
    \item Scenario 1: The robot navigates an intersection, encounters obstacles, and gets stuck in a narrow tunnel.
    \item Scenario 2: The robot hesitates at an intersection, explores a U-turn, then becomes stuck at a dead end.
    \item Scenario 3: After a U-turn and intersection, the robot fails to proceed into an unexplored tunnel.
    \item Scenario 4: The robot struggles with a slope, attempts again, but becomes stuck near an obstacle.
  
\end{itemize}

Baseline intervention timing is defined as the moment the robot reaches a failure point and remains stuck for 10 seconds (derived from scenario designer input). This controlled simulation ensures that every participant experiences an identical scene, eliminating extraneous variability (e.g., differences in lighting or object placement) that can occur in physical environments. Such consistency allows us to directly compare novice, intermediate, and expert intervention behaviors while still reflecting challenges derived from real-world scenarios.

\subsection{Participants Recruitment}

Twenty-seven participants (8 females, 19 males, age \(28.8 \pm 4.9\)) with diverse backgrounds were recruited via email invitations to staff and PhD students at CSIRO and at Monash University. The participants represented varying levels of HRT expertise, as defined in Section~\ref{subsec:expertise}. 

%Before the experiment, participants were asked to rate their familiarity with AI, HRI, and our simulator. Specifically, 11 participants had knowledge of AI but had never previously experienced interacting with robots. Another 11 participants were familiar with both AI and robotics but were new to this particular simulator and mission. Five participants had prior experience using the simulator. 

\subsection{Procedure}

Participants provided informed consent and reported their experience with human-robot collaboration via a questionnaire. They were seated in front of a laptop with a monitor displaying the simulated tunnel environment. Participants had a clear view of the exploration task interface and were comfortable with the input devices provided. Before interacting with the simulated robot, they watched a video explaining the steps of this user study, the task of the robot and participant, and the interface. A practice session followed, allowing them to familiarize themselves with the interface and how to place waypoints. Participants then completed two quizzes: one to confirm task understanding (with feedback) and another to assess expertise (without feedback). Each participant participated in all 4 scenarios (Fig.~\ref{fig:map}). After each scenario, participants were asked to rate their intervention timing (timely, early, or late) and explain the reasons behind their interventions. A post-experiment questionnaire captured their overall experience and intervention preferences.
    %c) Pre-Experiment Introduction Training Video: Create an informative training video to introduce participants to the study's purpose, procedures, and the simulated tunnel environment. The video will cover essential aspects, including the robot's goal, the participant's role, and how to provide waypoints during the exploration task.
    %d) Pre-recorded scenarios simulating robot troubles stored in ROS bags: Create scenarios where the robot encounters various challenges or malfunctions, such as navigation errors or exploring an area that has already been explored. These scenarios are recorded and stored in ROS (Robot Operating System) bags, a file format used to capture data streams in robotics, enabling the replay of scenarios during user studies or system testing.
    %e) Participant Preparation: Participants will be seated in front of a laptop with a monitor displaying the simulated tunnel environment. We will ensure participants have a clear view of the exploration task interface and are comfortable with the input devices provided.
    %f) Exploration Task Overview: We will clearly communicate the objective of the robot: to explore as much of the simulated tunnel environment as possible within a limited timeframe. We will emphasize that participants play a crucial role by providing waypoints to the robot when it encounters challenges or attempts to re-explore previously covered areas.

\subsection{Experimental Measures}
We collected several measures during and after the simulator experiments. These measures were designed to evaluate various aspects of the interaction, including intervention behaviours, robot performance, and user satisfaction.

\subsubsection{Task-related measures}
% We collected the following task-related measures during the participant interactions with the robot:

\textbf{Intervention timing:} Intervention timing refers to the specific moment during the task when participants chose to intervene and provide assistance to the robot. This metric is essential for understanding the decision-making process of human operators, as it reveals whether participants tend to intervene early as soon as they perceive the robot encountering an issue or late after the robot has already been grappling with a problem. Recorded as timestamps corresponding to each intervention, this metric enables analysis of intervention intervals and their correlation with task performance.

\textbf{Waypoint position:} Waypoint position refers to the specific locations on the map where participants set waypoints to guide the robot. This metric provides insights into the spatial strategies employed by operators. Effective waypoint positioning is crucial for efficient navigation and task completion, as optimal waypoints help the robot avoid obstacles and take the shortest or safest path. Analyzing waypoint positions highlights strategic differences across expertise levels. Recorded as coordinates and orientations, they reveal variations in distribution, obstacle proximity, and path alignment, reflecting differences in planning and effectiveness.

\textbf{Area Covered Post-Intervention:} Area covered post-intervention refers to the total area explored by the robot after a human intervention has occurred. The interface in the user study is not fully interactive. We terminated the simulation once the participant provided a waypoint. To examine the performance of the intervention, we subsequently run the simulation forward to see what would have happened as a result of that intervention. The metric captures the effectiveness of the intervention by measuring how much area the robot successfully explores after the operator intervened compared to autonomous behaviours without the intervention.

\subsubsection{Questionnaire measures} 
We collected questionnaire measures to capture participants' subjective experiences and perceptions during the experiment. The questions focused on \textbf{task evaluation},\textbf{ self-evaluation}, and \textbf{interaction preferences}. Participants were asked to rate task difficulty (1 = Not challenging at all, 10 = Extremely challenging), overall satisfaction (1 = Not satisfied at all, 10 = Extremely satisfied), and confidence in their interventions (1 = Not confident at all, 10 = Extremely confident). To evaluate interaction preferences, participants indicated whether they would assist if the robot explicitly asked for help and whether they preferred assistance to be initiated by the robot or at their own discretion. They were also asked to answer open-ended questions related to \textbf{system performance} and \textbf{overall experience}. These responses provided insights into how expertise levels influenced task perception, confidence, and help-seeking behavior.

\section{Results Analysis}

We analyzed intervention strategies across expertise levels using quantitative and qualitative data. Before conducting ANOVA, normality and homogeneity of variance assumptions were validated via Shapiro-Wilk (\(p > 0.05\) for all groups) and Levene’s tests (\(p > 0.05\)), respectively. 

\subsection{Level of Expertise}\label{subsec:expertise}

We categorized participants as three expertise levels: Novice, Intermediate, and Experienced, based on their background and experience. This classification was conducted post-study by reviewing self-reported survey on familiarity with AI, robotics, and similar HRT task or simulation.
\begin{itemize}
 
    \item \textbf{Novice:} Participants (N=11) with little or no experience in AI, robotics, or relevant simulations.
    \item \textbf{Intermediate:} Participants (N=11) with moderate experience or knowledge in AI and robotics but new to this specific simulator and mission.
    \item \textbf{Experienced:} Participants (N=5) with substantial experience in using the simulator or in AI and robotics.
    
\end{itemize}

%Specifically, participants without previous experience in our simulator were asked to rate their familiarity with robotics and AI technologies. 3 participants rate their familiarity as ``Low''; 8 participants rated themselves themselves as ``Moderate'', 3 participants rated it as ``High''; 1 participant rated it as ``Very High''. Those who rated their familiarity ``Low'' had no prior experience interacting with any robotic system. Similarly, 4 participants who rated their familiarity as ``Moderate'' also had no prior experience with robotics. The remaining participants had experience in HRI. 
Experienced participants had between one to five years of simulator experience, with usage frequency ranging from monthly to daily. Given the specialized nature of this domain and the proprietary NavStack simulator, expert operators are scarce, limiting the participant pool. 

\begin{table*}[ht]
    \centering
    \caption{Mixed-design ANOVA results of intervention behaviors: Shapiro-Wilk ($p > 0.05$ for all groups) and Levene’s tests ($p > 0.05$); Significance code: $0<***<0.001<**<0.01<*<0.05$}
    \begin{tabular}{llcccccc}
        % Shapiro-Wilk ($p > 0.05$ for all groups) and Levene’s tests ($p > 0.05$)\\
        \toprule        
        Data & Variables & Sum of Squares & df & F & p & $\eta^2_{p}$ \\
        \midrule
        \multirow{4}{*}{Intervention Timing} 
        & C(scenario) & $5.29e+05$ & 3 & 14.6 & $6.39e-08^{***}$ & 0.31 \\
        & C(expertise) & $1.29e+05$ & 2 & 5.35 & $6.22e-03^{**}$ & 0.10 \\
        & C(scenario):C(expertise) & $2.92e+04$ & 6 & 0.4 & $8.75e-01$ & 0.025 \\
        & Residual & $1.10e+06$ & 96 & - & - & - \\
        \midrule
        \multirow{4}{*}{Intervention Waypoint} 
        & C(scenario) & 1597.33 & 3 & 35.97 & $1.14e-15^{***}$ & 0.268 \\
        & C(expertise) & 334.73 & 2 & 3.40 & $3.73e-02^{*}$ & 0.071 \\
        & C(scenario):C(expertise) & 1116.29 & 6 & 0.12 & $9.9e-01$ & 0.004 \\
        & Residual & 4356.90 & 96 & - & - & - \\
        \midrule
        \multirow{4}{*}{Area Covered Post-Intervention} 
        & C(scenario) & $2.75e+06$ & 3 & 38.42 & $2.12e-16^{***}$ & 0.546 \\
        & C(group) & $2.14e+05$ & 3 & 2.97 & $3.55e-02^{*}$ & 0.085 \\
        & C(scenario):C(group) & $1.92e+05$ & 9 & 0.89 & $5.36e-01$ & 0.077 \\
        & Residual & $3.51e+08$ & 96 & - & - & - \\
        \bottomrule
        % Significance code: $0<***<0.001<**<0.01<*<0.05$
    \end{tabular}
    \label{tab:anova_results}
\end{table*}

\subsection{Operator Expertise and Intervention Behaviours - \textbf{RQ1}}

\subsubsection{Intervention Timing by Expertise Level}

We ran a mixed-design ANOVA with scenarios (4 scenarios) and expertise level (Novice vs Intermediate vs Experienced) as independent variables (Table \ref{tab:anova_results}). Our analysis revealed significant main effects of the type of scenario (\(F(3, 96) = 14.60, p =6.39e-08 < 0.001, \eta^{2}_{p}=0.31\)) and expertise (\(F(2, 96) = 5.35, p = 6.22e-03 < 0.01, \eta^{2}_{p}=0.10\)) on the intervention timing, but no interaction effect \(F(6, 96) = 0.40, p = 8.75e-01, \eta^{2}_{p}=0.025\). Tukey HSD revealed Scenario 3 differs significantly from others (\(p<0.001\)). Significant differences are observed between Intermediate and Novice users (\(p=1.6e-03<0.005\)), and between Experts and Novices (\(p=2.80e-03<0.005\))

\begin{figure}
    \centering
    \includegraphics[width=0.8\linewidth]{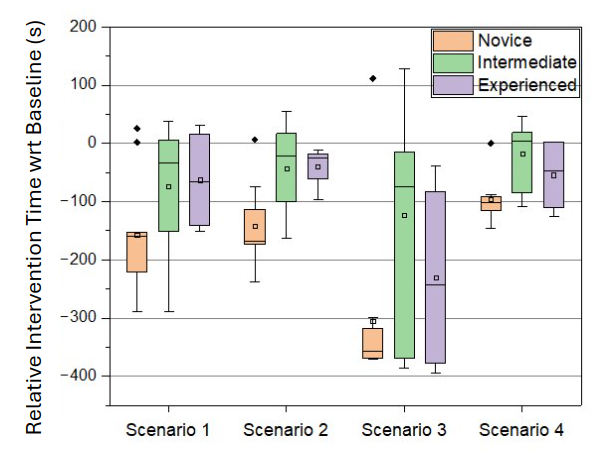}
    \caption{Response Times by Scenario and Group}
    \label{fig:results_time}
\end{figure}

Fig. \ref{fig:results_time} reveals distinct patterns of intervention timing among different expertise levels (addressing
\textbf{RQ1.1}). The relative response time is computed as the participant's intervention timing minus the baseline (i.e., 0s on the y-axis is the pre-defined robot failure point in each map). Novices intervene significantly earlier than intermediate and experienced users, with lower variance in timing. Participant reasoning (see Section \ref{sec:feedback}) provides insights into this trend — Novices rely on clear cues like stalling, obstacle proximity, or revisiting areas (e.g., 'The robot was stuck near a wall.' - P3, Novice, Scenario 1), leading to lower variance in intervention timing. In contrast, Intermediate and Expert users intervene later, closer to predefined failure points. Their familiarity with the robot’s behaviour enables them to give it more time to resolve issues autonomously before stepping in. Similarly, Lewis et al.~\cite{lewis2013adjustable} found that Novices issued stop commands more frequently, while Experts exercised more decisive control. Both studies suggest that Novices take a more conservative approach, while Experts intervene more strategically in remote HRT (answering \textbf{RQ1}).

\subsubsection{Intervention Waypoint by Expertise Level}

A mixed-design ANOVA (Table \ref{tab:anova_results}) found significant main effects of scenario type (\(F(3,96)=35.97, p=1.14e-15 <0.001, \eta^{2}_{p}=0.268\)), and expertise level (\(F(2,96)=3.40, p= 3.73e-02<0.05, \eta^{2}_{p}=0.071\)) on intervention waypoint positions, but no significant interaction effect (\(F(6,96)=0.12, p= 9.9e-01, \eta^{2}_{p} =0.004\)). Post-hoc Tukey HSD tests confirmed that waypoint placement significantly differed between all scenario pairs (\(p<0.01\)). Significant differences are observed between expert and novice users (\(p=0.003<0.005\)). 

\begin{figure*}
    \centering
    \includegraphics[width=0.6\linewidth]{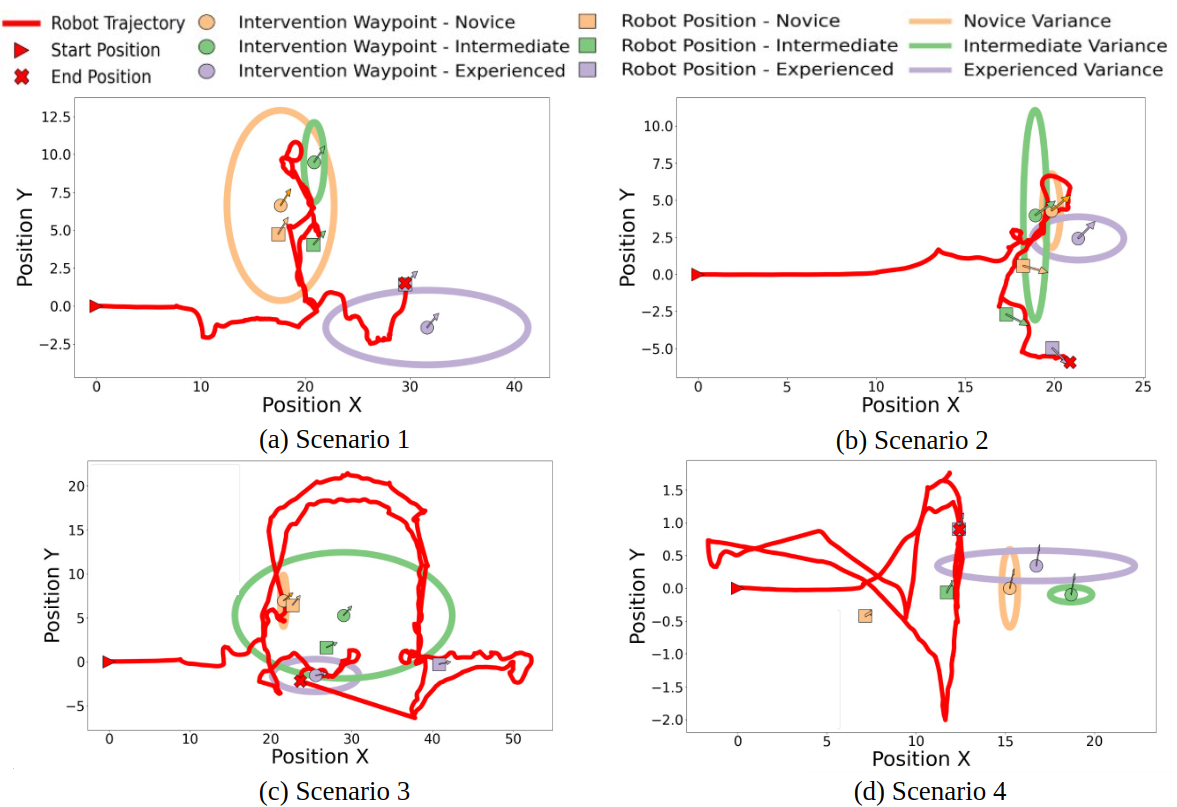}
    \caption{User intervention relating to robot trajectory. Waypoints by Novices, Intermediates, and Experts are marked as orange, green, and purple circles, respectively, with arrows indicating mean orientation. Corresponding robot positions are shown as squares in the same colors. Variance ellipses depict the spread of intervention points. The robot's trajectory is plotted in red (arrow for start and X for end). }
    \label{fig:trajec}
\end{figure*}

To answer \textbf{RQ1.2}, we examined Fig.~\ref{fig:trajec}, which illustrates the robot's trajectory along with the mean positions and orientations of waypoints placed by participants and the corresponding robot positions at those times. Novices intervene early, while intermediates and experts wait until the robot encounters obstacles before intervening, aligning with findings from Fig.~\ref{fig:results_time}. 

The intervention positions of operators varied by expertise level across scenarios, as depicted in Fig. \ref{fig:trajec}. Most participants place waypoints near the robot’s current location, regardless of expertise. Experts, however, show a consistent intervention strategy, with low variance in placement. Beyond placement, waypoint orientation is key to success. Experts guide the robot away from failure points, while novices and intermediates often misalign waypoints, causing inefficient recovery and unnecessary detours. In Scenario 1, novices and intermediates direct the robot into a narrow tunnel, failing to recognize the challenge.

%\begin{figure*}
    %\centering
	  %\subfloat[Scenario 1]{
       %\includegraphics[width=7cm,keepaspectratio]{Fig7.png}}
    %\label{marker_set_map1}
        %\subfloat[Scenario 2]{
       %\includegraphics[width=7cm,keepaspectratio]{Fig8.png}}
    %\label{marker_set_ map2}
         %\subfloat[Scenario 3]{
       %\includegraphics[width=7cm,keepaspectratio]{Fig9.png}}
    %\label{marker_set_ map3}
         %\subfloat[Scenario 4]{
       %\includegraphics[width=7cm,keepaspectratio]{Fig10.png}}
    %\label{marker_set_ map4}

    %\caption{Mean Positions and Orientations of User Group Interventions and Robot Trajectory. The mean positions of the intervention waypoint provided by novice users, intermediate users, and experts are marked with orange, green, and purple circles, respectively, each with an orientation arrow indicating the mean orientation. Similarly, the positions of the robot during these interventions are represented by squares in the same colors (orange for novice users, green for intermediate users, and purple for experts), each also accompanied by an orientation arrow. Variance ellipses are overlaid on the mean positions to indicate the spread of intervention points for each user group.The robot's trajectory is plotted in red, showing its path over time. The start and end positions of the robot are highlighted with a red arrow and a red X marker, respectively. }
    \label{fig:tra} 
  
%\end{figure*}

\subsubsection{Area Covered Post-Intervention}
\label{sec:area_post}
To further study \textbf{RQ1.2}, a forward simulation was conducted to examine the effectiveness of participant interventions. A mixed-design ANOVA results (Table \ref{tab:anova_results}) show significant main effects of the type of scenario (\(F(3,96)=38.42, p=2.12e-16<0.001, \eta^{2}_{p}=0.546 \)) and expertise group (Novice, Intermediate, Experienced, and Robot) (\(F(3,96)=2.97, p=3.55e-02, \eta^{2}_{p}=0.085\)) on the area covered, with no interaction effect (\(F(9,96)=0.89, p=5.36e-01, \eta^{2}_{p} =0.077\) ). Post hoc Tukey’s HSD tests confirmed significant differences across all scenario comparisons (\(p < 0.001\)), with Scenario 3 deviating most from the others. For expertise groups, Experts and Intermediates significantly outperformed novices and robots (\(p<0.001\)), but the difference between Experts and Intermediates was marginal (\(p=0.05\)). Area covered following intervention by Novices compared to robots operating alone showed no significant difference (\(p=0.13\)).

\begin{figure*}
    \centering
    \includegraphics[width=0.6\linewidth]{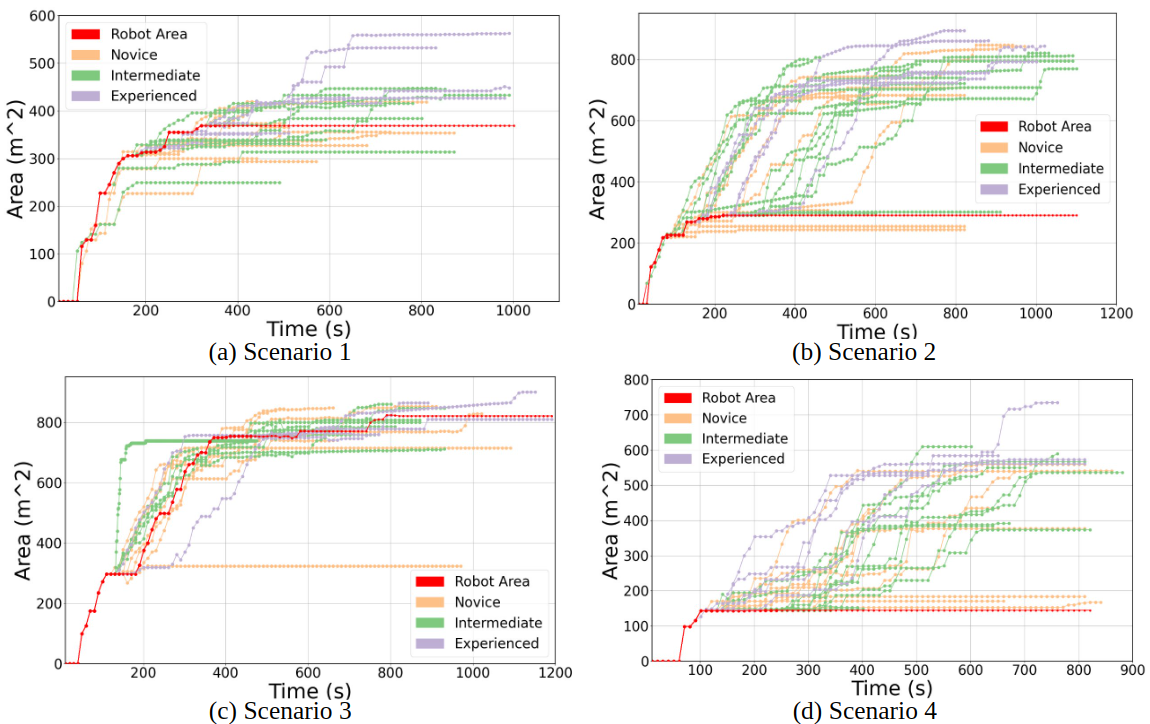}
    \caption{Area covered vs. time for the expertise group and the
autonomous robot baseline. The plot compares area coverage over time for each novice, intermediate, and experienced users, along with the autonomous robot baseline. }
    \label{fig:forward}
\end{figure*}

\begin{figure*}
    \centering
	  \subfloat[Scenario 1\label{fig:sc1t}]{
       \includegraphics[width=4.4cm,keepaspectratio]{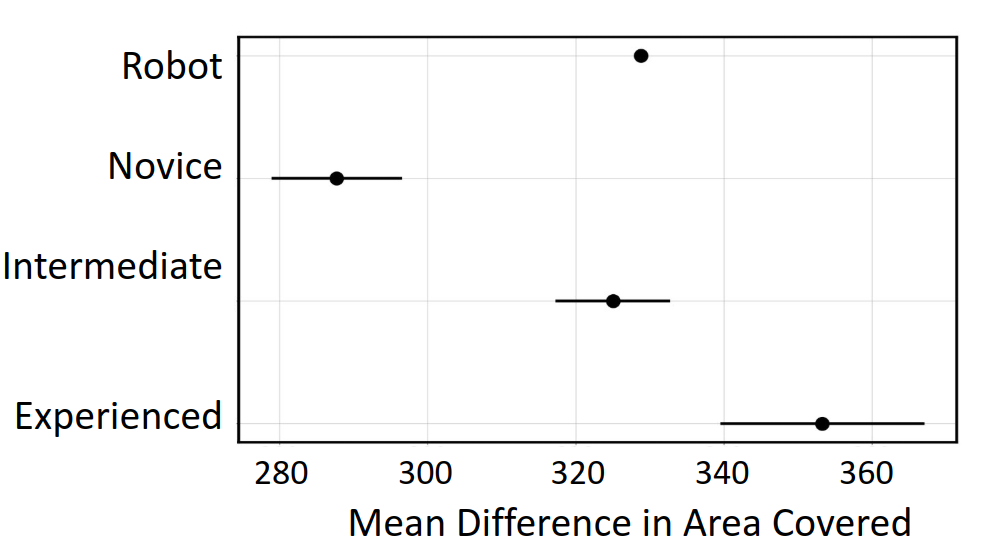}}
        \subfloat[Scenario 2\label{fig:sc2t}]{
       \includegraphics[width=4.4cm,keepaspectratio]{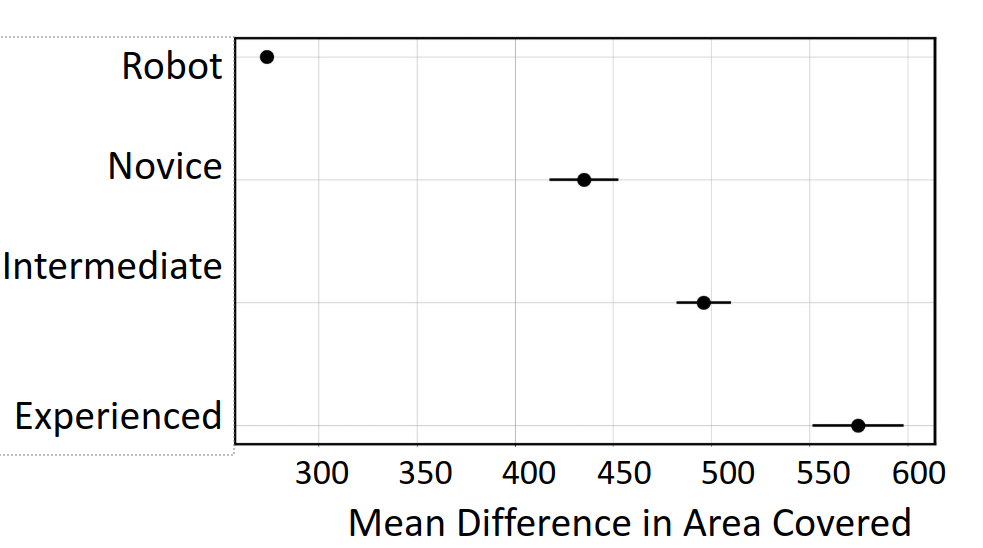}}
         \subfloat[Scenario 3\label{fig:sc3t}]{
       \includegraphics[width=4.4cm,keepaspectratio]{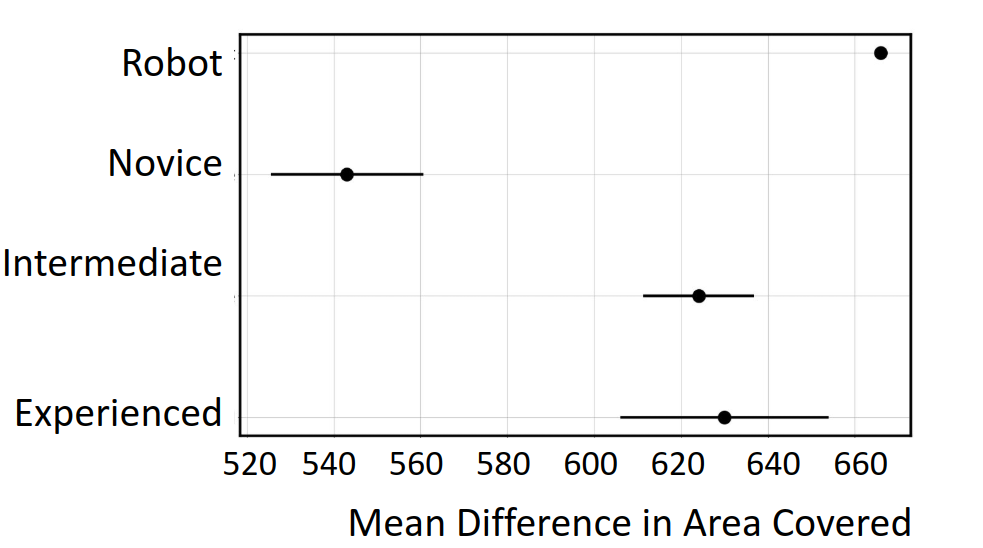}}
         \subfloat[Scenario 4\label{fig:sc4t}]{
       \includegraphics[width=4.4cm,keepaspectratio]{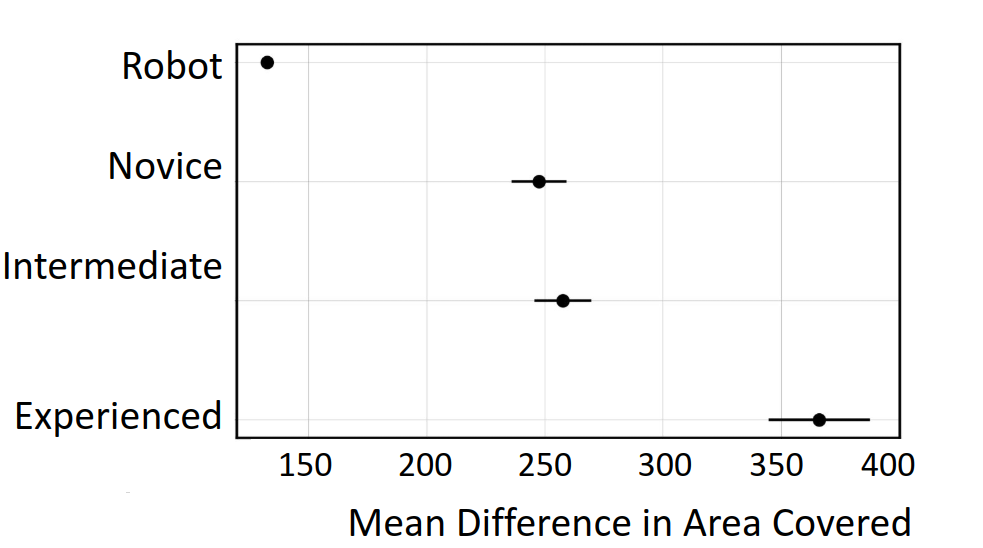}}   
 
    \caption{Mean differences in area covered by groups. The x-axis shows the area covered. The y-axis lists Robot, Novice, Intermediate, and Experienced groups. Horizontal lines indicate confidence intervals, with central dots being mean values.}
    \label{fig:hsd} 
  
\end{figure*}

Fig. \ref{fig:forward} shows the area covered by the robot before and after intervention, while Fig. \ref{fig:hsd} presents confidence intervals for each group's mean area covered. Human intervention generally improves performance over autonomy in scenarios 1, 2, and 4, with Experienced operators achieving the highest coverage. However, variability among Novice and Intermediate participants suggests that not all interventions are effective. Intermediate participants initially outperformed the autonomous robot but plateaued, while some Novice interventions even reduced performance (Fig. \ref{fig:forward}).

In scenario 3, the robot’s autonomous performance surpasses that of human intervention. In this scenario, the robot did not encounter significant challenges or become stuck until a critical point, where it chose to explore a longer, already traversed tunnel instead of a shorter, unexplored area at an intersection, which occurred much later close to mission completion (see Section \ref{sec:sc_design}). Only a few participants recognized and corrected this behavior. This suggests that under specific conditions, such as tasks closely aligned with the robot's programming, the robot’s autonomous exploration performance can be more advantageous.

%\begin{figure*}
    %\centering
	  %\subfloat[Scenario 1]{
       %\includegraphics[width=6cm,keepaspectratio]{Fig7.png}}
        %\subfloat[Scenario 2\label{fig:sc2f}]{\includegraphics[width=6cm,keepaspectratio]{Fig8.png}}

         %\subfloat[Scenario 3]{
       %\includegraphics[width=6cm,keepaspectratio]{Fig9.png}}
         %\subfloat[Scenario 4]{
       %\includegraphics[width=6cm,keepaspectratio]{Fig10.png}}

    %\caption{Area covered vs. time for participant groups and robot performance. The plot compares area coverage over time for each novice, intermediate, and experienced users, along with the autonomous robot baseline.}
        
    %\label{fig:forward} 
  
%\end{figure*}

\subsection{Confidence in Intervention}
\label{sec:confident}
Confidence ratings were collected from 19 of 27 participants (5 Experts, 7 Intermediates, 7 Novices); data from 8 were missing due to incomplete responses. Experts reported high and consistent confidence (\(M = 6.8, SD = 1.10\)). Intermediates had the lowest confidence (\(M = 4.17, SD = 1.47\)), showing greater uncertainty and variability. Novices reported the highest confidence (\(M = 7.0, SD = 1.41\)), but their variability suggests less consistent self-assessment. A one-way ANOVA confirmed a significant effect of expertise level on confidence (\(F(2,16)=23.29,p<0.001\)). Post-hoc Tukey’s HSD tests further showed that intermediates had significantly lower confidence than both Experts  (\(p=0.007\)) and Novices (\(p=0.003\)), while no significant difference was found between Experts and Novices(\(p=0.97\)). These results suggest that confidence does not increase linearly with expertise. Novices may overestimate their abilities, while Intermediates are more aware of their limitations.

This aligns with participant reasoning in Section \ref{sec:feedback}, where Novices initially hesitated but gradually gained confidence in their decisions on when to intervene. However, despite their confidence, Novice interventions performed worse than those of Experienced participants and even underperformed compared to the robot's autonomous exploration in some scenarios (see Section \ref{sec:area_post}). These findings align with~\cite{conlon2022m}, which highlights the tendency for overconfidence in Novice operators.

\subsection{Participants' Intervention Strategies}
\label{sec:feedback}
After each scenario, we asked each participant to give observations and reasons for their intervention behaviours. Based on what they could observe, participant interventions likely relied on the following factors:

\textbf{Robot Behaviour:} Most participants observed the robot to decide when to intervene. Novices relied on clear signs like the robot stopping or failing to progress. For instance, one noted, “Since it did not move for some time, I could have intervened quicker.” (P5, Novice, Scenario 1). In contrast, experts allowed more time for autonomous operation and intervened only upon detecting inefficiencies or failures. “Provided enough time for the robot to explore autonomously but intervened due to inefficient navigation.” (P22, Experienced, Scenario 1). Another expert emphasized patience, noting they intervened for efficiency but recognized the system might have self-corrected: “If I had been more patient, the system looked like it would have reverted to the most efficient task.” (P22, Experienced, Scenario 2). Some novices adapted their approach over time, shifting towards efficiency-based interventions: “As soon as I saw continued difficulty, I intervened to redirect it.” (P5, Novice, Scenario 2).

\textbf{Timing Based Responses:} Timing played a crucial role for many participants. ``I waited 10-20 seconds when the robot became stuck against a wall before intervening, which I think is reasonable as it sometimes figures out an issue after that amount of time.'' (P14, Intermediate, Scenario 3). This participant used a timing rule, allowing the robot some time to resolve the issue before stepping in. 

\textbf{Environmental Cues:} Participants frequently recognized environmental factors, such as walls, narrow passages, or dead ends, as key indicators that the robot required intervention. These environmental cues often triggered decisions to intervene when the robot struggled to navigate or became stuck. For instance, one participant noted, ``The robot has been stuck by an obstacle and returned to the travelled route.'' (P16, Novice, Scenario 1). Similarly, another participant observed, ``The robot was stuck near a wall.'' (P3, Novice, Scenario 1). These participants responded to clear physical barriers that visibly hindered the robot's progress.

\subsection{``Ask for help'' Preferences - \textbf{RQ2}}
To answer \textbf{RQ2}, participants were asked about their interaction preferences after the experiment. In the post-experiment survey, all participants responded ``Yes'' to the question, ``If the robot asks for help, would you be willing to provide assistance?'' except for two individuals who expressed no preference. 
% Fig.~\ref{fig:results_preference} shows the number of participants in response to 
In the subsequent question, ``How do you prefer that help be initiated?'', a chi-square test confirmed a statistically significant difference in help initiation preferences (\(\chi^2 (2, N=27)= 12.67, p = 0.0018\)), though expertise level had no significant effect.
The majority of participants expressed a preference for the robot explicitly asking for help, a trend that was particularly pronounced among novice operators. This finding suggests that less experienced users prefer that the robot provides clear cues for when assistance is needed. In contrast, participants with higher levels of experience were more likely to intervene and assist the robot without explicit requests. These responses indicate that experienced operators are more proactive and confident in their ability to identify when the robot requires help, leading to more efficient intervention behaviours.

% \begin{figure}
%     \centering
%     \includegraphics[width=0.7\linewidth]{Fig16.PNG}
%     \caption{Number of participants in response to the question ``How do you prefer that help be initiated?''}
%     \label{fig:results_preference}
% \end{figure}

\section{Discussion}
This study examined how operator expertise influences intervention behaviours, intervention timing, and help-seeking preferences in remote human-robot teaming. \textbf{Expertise influences intervention timing and effectiveness.}
Novices intervened earlier and more consistently, relying on obvious visual cues (e.g., stalling, obstacles) but often overestimated their abilities (Section \ref{sec:confident}). Experts delayed intervention, allowing for robot self-recovery, which enhanced efficiency. These differences are supported by statistically significant results (Table \ref{tab:anova_results}).
\textbf{Human intervention generally improves performance—but not always.}
Human intervention generally improved performance (Scenarios 1, 2, and 4), especially among experienced users. However, Scenario 3 revealed that autonomous navigation outperformed human intervention, suggesting that robot autonomy can sometimes be more effective when well-aligned with task demands (Section \ref{sec:sc_design}). These results, consistent with Robinson et al.~\cite{robinson2022human}, show that our simulator captures the variability of human performance observed in real-world settings while isolating key decision points.
\textbf{Most participants preferred robot-initiated help requests}, with novices relying on explicit cues, while experts preferred self-directed intervention. This suggests that novices need clearer guidance, whereas experts are more confident in assessing intervention needs. Future interfaces should adapt help prompts based on expertise, providing structured guidance for novices while minimizing interruptions for experts. Instead of trying to train all operators to behave like experts, adaptive robots could use real-time behavioural signals to estimate operator expertise and adjust their autonomy and communication accordingly. %Despite limitations, this study offers insights into how expertise shapes intervention strategies in remote human–robot teams. 
\textbf{This study has several limitations that should be considered }when interpreting the results. The expert group was small (\(n = 5\)) due to the difficulty of recruiting operators with substantial robotic exploration experience, but we included all available experts to capture their behaviours. The single-intervention-per-scenario design was chosen to control variability and isolate the decision-making process for the first intervention, though it limits the natural expression of strategies. Although the simulated tunnels use realistic maps, they cannot fully capture field uncertainties. Future work will expand the expert pool, allow repeated interventions to observe the evolution of behavioural and mental models, refine expertise measures, and validate findings in field trials to inform adaptive shared autonomy.

%Finally, while the simulated tunnel environments were based on realistic layouts, they cannot fully reproduce the uncertainties of field operations. Future work should enlarge the expert pool, allow repeated interventions to observe the evolution of behaviours and mental models, refine expertise measures, and validate the findings in field trials. These extensions would support developing shared autonomy systems that can recognise and adapt to operators’ expertise in real time, improving collaboration without imposing additional training burdens. %Despite these constraints, this study offers valuable insights into how expertise shapes intervention strategies in remote human–robot teams.

\section{CONCLUSIONS}

This study examined how operator expertise influences intervention behavior in simulated missions, offering insights to enhance human-robot collaboration. Conducted in a high-fidelity simulator with realistic maps and robots, our findings highlight ways to lower entry barriers for non-roboticists while improving task success. By understanding how operator expertise affects intervention behaviour, we can inform the design of adaptive shared autonomy systems that collaborate effectively with users of diverse expertise levels.

\section*{Acknowledgments}
This project was funded by CSIRO’s CINTEL FSP program.

%% This section was initially prepared using BibTeX.  The .bbl file was
%% placed here later
\bibliography{IEEEexample}

@article{vasconez2019human,
  title={Human--robot interaction in agriculture: A survey and current challenges},
  author={Vasconez, Juan P and Kantor, George A and Cheein, Fernando A Auat},
  journal={Biosystems engineering},
  volume={179},
  pages={35--48},
  year={2019},
  publisher={Elsevier}
}

@article{ajoudani2018progress,
  title={Progress and prospects of the human--robot collaboration},
  author={Ajoudani, Arash and Zanchettin, Andrea Maria and Ivaldi, Serena and Albu-Sch{\"a}ffer, Alin and Kosuge, Kazuhiro and Khatib, Oussama},
  journal={Autonomous Robots},
  volume={42},
  pages={957--975},
  year={2018},
  publisher={Springer}
}

@article{nourbakhsh2005human,
  title={Human-robot teaming for search and rescue},
  author={Nourbakhsh, Illah R and Sycara, Katia and Koes, Mary and Yong, Mark and Lewis, Michael and Burion, Steve},
  journal={IEEE Pervasive Computing},
  volume={4},
  number={1},
  pages={72--79},
  year={2005},
  publisher={IEEE}
}

@inproceedings{conlon2022m,
  title={“I'm Confident This Will End Poorly”: Robot Proficiency Self-Assessment in Human-Robot Teaming},
  author={Conlon, Nicholas and Szafir, Daniel and Ahmed, Nisar},
  booktitle={IROS'22},
  pages={2127--2134},
  year={2022},
  organization={IEEE}
}

@article{robinson2022human,
  title={Human-Robot Team Performance Compared to Full Robot Autonomy in 16 Real-World Search and Rescue Missions: Adaptation of the DARPA Subterranean Challenge},
  author={Robinson, Nicole and Williams, Jason and Howard, David and Tidd, Brendan and Talbot, Fletcher and Wood, Brett and Pitt, Alex and Kottege, Navinda and Kuli{\'c}, Dana},
  journal={arXiv preprint arXiv:2212.05626},
  year={2022}
}

@article{othman2023human,
  title={Human--robot collaborations in smart manufacturing environments: review and outlook},
  author={Othman, Uqba and Yang, Erfu},
  journal={Sensors},
  volume={23},
  number={12},
  pages={5663},
  year={2023},
  publisher={MDPI}
}

@article{hopko2022human,
  title={Human factors considerations and metrics in shared space human-robot collaboration: A systematic review},
  author={Hopko, Sarah and Wang, Jingkun and Mehta, Ranjana},
  journal={Frontiers in Robotics and AI},
  volume={9},
  pages={799522},
  year={2022},
  publisher={Frontiers Media SA}
}

@article{wijayathunga2023challenges,
  title={Challenges and solutions for autonomous ground robot scene understanding and navigation in unstructured outdoor environments: A review},
  author={Wijayathunga, Liyana and Rassau, Alexander and Chai, Douglas},
  journal={Applied Sciences},
  volume={13},
  number={17},
  pages={9877},
  year={2023},
  publisher={MDPI}
}

@article{cunha2022exploring,
  title={Exploring the status of the human operator in Industry 4.0: A systematic review},
  author={Cunha, Liliana and Silva, Daniel and Maggioli, Sarah},
  journal={Frontiers in Psychology},
  volume={13},
  pages={889129},
  year={2022},
  publisher={Frontiers Media SA}
}

@article{hambuchen2021review,
  title={A review of NASA human-robot interaction in space},
  author={Hambuchen, Kimberly and Marquez, Jessica and Fong, Terrence},
  journal={Current Robotics Reports},
  volume={2},
  number={3},
  pages={265--272},
  year={2021},
  publisher={Springer}
}

@inproceedings{riley2004hunt,
  title={The hunt for situation awareness: Human-robot interaction in search and rescue},
  author={Riley, Jennifer M and Endsley, Mica R},
  booktitle={Proc. Hum. Factors Ergon. Soc. Annu. Meet.},
  volume={48},
  number={3},
  pages={693--697},
  year={2004},
  organization={SAGE}
}

@article{li2023proactive,
  title={Proactive human--robot collaboration: Mutual-cognitive, predictable, and self-organising perspectives},
  author={Li, Shufei and Zheng, Pai and Liu, Sichao and Wang, Zuoxu and Wang, Xi Vincent and Zheng, Lianyu and Wang, Lihui},
  journal={Robotics and Computer-Integrated Manufacturing},
  volume={81},
  pages={102510},
  year={2023},
  publisher={Elsevier}
}

@article{chen2014human,
  title={Human--agent teaming for multirobot control: A review of human factors issues},
  author={Chen, Jessie YC and Barnes, Michael J},
  journal={IEEE Trans. Hum. Mach. Syst.},
  volume={44},
  number={1},
  pages={13--29},
  year={2014},
  publisher={IEEE}
}

@article{burke2004moonlight,
  title={Moonlight in Miami: Field study of human-robot interaction in the context of an urban search and rescue disaster response training exercise},
  author={Burke, Jennifer L and Murphy, Robin R and Coovert, Michael D and Riddle, Dawn L},
  journal={HCI},
  volume={19},
  number={1-2},
  pages={85--116},
  year={2004},
  publisher={Taylor \& Francis}
}

@inproceedings{scholtz2004evaluation,
  title={Evaluation of human-robot interaction awareness in search and rescue},
  author={Scholtz, Jean and Young, Jeff and Drury, Jill L and Yanco, Holly A},
  booktitle={ICRA'04},
  volume={3},
  pages={2327--2332},
  year={2004},
  organization={IEEE}
}

@inproceedings{wang2016trust,
  title={Trust calibration within a human-robot team: Comparing automatically generated explanations},
  author={Wang, Ning and Pynadath, David V and Hill, Susan G},
  booktitle={HRI'16},
  pages={109--116},
  year={2016},
  organization={IEEE}
}

@article{sheridan2016human,
  title={Human--robot interaction: status and challenges},
  author={Sheridan, Thomas B},
  journal={Human factors},
  volume={58},
  number={4},
  pages={525--532},
  year={2016},
  publisher={SAGE}
}

@article{kottege2023heterogeneous,
  title={Heterogeneous robot teams with unified perception and autonomy: How Team CSIRO Data61 tied for the top score at the DARPA Subterranean Challenge},
  author={Kottege, Navinda and Williams, Jason and Tidd, Brendan and Talbot, Fletcher and Steindl, Ryan and Cox, Mark and Frousheger, Dennis and Hines, Thomas and Pitt, Alex and Tam, Benjamin and others},
  journal={arXiv preprint arXiv:2302.13230},
  year={2023}
}

@article{ramezani2022wildcat,
  title={Wildcat: Online continuous-time 3d lidar-inertial slam},
  author={Ramezani, Milad and Khosoussi, Kasra and Catt, Gavin and Moghadam, Peyman and Williams, Jason and Borges, Paulo and Pauling, Fred and Kottege, Navinda},
  journal={arXiv preprint arXiv:2205.12595},
  year={2022}
}

@article{licardo2024intelligent,
  title={Intelligent robotics—A systematic review of emerging technologies and trends},
  author={Licardo, Josip Tomo and Domjan, Mihael and Orehova{\v{c}}ki, Tihomir},
  journal={Electronics},
  volume={13},
  number={3},
  pages={542},
  year={2024},
  publisher={MDPI}
}

@inproceedings{lewis2013adjustable,
  title={An adjustable autonomy paradigm for adapting to expert-novice differences},
  author={Lewis, Bennie and Tastan, Bulent and Sukthankar, Gita},
  booktitle={IROS'13},
  pages={1656--1662},
  year={2013},
  organization={IEEE}
}

@article{sfair2020human,
  title={Human--robot interface for embedding sliding adjustable autonomy methods},
  author={Sfair Palar, Piatan and de Vargas Terres, Vin{\'\i}cius and Schneider de Oliveira, Andr{\'e}},
  journal={Sensors},
  volume={20},
  number={20},
  pages={5960},
  year={2020},
  publisher={MDPI}
}

@inproceedings{storms2014predicting,
  title={Predicting human performance during teleoperation},
  author={Storms, Justin and Vozar, Steven and Tilbury, Dawn},
  booktitle={HRI'14},
  pages={298--299},
  year={2014}
}

@article{liang2024learning,
  title={Learning to learn faster from human feedback with language model predictive control},
  author={Liang, Jacky and Xia, Fei and Yu, Wenhao and Zeng, Andy and Arenas, Montserrat Gonzalez and Attarian, Maria and Bauza, Maria and Bennice, Matthew and Bewley, Alex and Dostmohamed, Adil and others},
  journal={arXiv preprint arXiv:2402.11450},
  year={2024}
}

@inproceedings{cakmak2014teaching,
  title={Teaching people how to teach robots: The effect of instructional materials and dialog design},
  author={Cakmak, Maya and Takayama, Leila},
  booktitle={HRI'14},
  pages={431--438},
  year={2014}
}

@inproceedings{sakr2020training,
  title={Training human teacher to improve robot learning from demonstration: A pilot study on kinesthetic teaching},
  author={Sakr, Maram and Freeman, Martin and Van der Loos, HF Machiel and Croft, Elizabeth},
  booktitle={RO-MAN'20},
  pages={800--806},
  year={2020},
  organization={IEEE}
}
\bibliographystyle{named}
%% The file named.bst is a bibliography style file for BibTeX 0.99c

\end{document}